\documentclass{article}

\usepackage{PRIMEarxiv}

\usepackage[sort&compress,square,comma,authoryear]{natbib}
\usepackage[utf8]{inputenc} %
\usepackage[T1]{fontenc}    %
\usepackage{hyperref}       %
\usepackage{url}            %
\usepackage{booktabs}       %
\usepackage{amsfonts}       %
\usepackage{nicefrac}       %
\usepackage{microtype}      %

\usepackage{algpseudocode}
\usepackage{algorithm}
\usepackage{svg}
\usepackage{caption}
\usepackage{subcaption}
\usepackage{amsmath}
\usepackage{amssymb}
\usepackage{placeins}
\usepackage{multirow}
\usepackage{cleveref}
\usepackage{xcolor}

\title{DARTFormer: Finding The Best Type Of Attention}

\author{Jason Ross Brown\\
University of Cambridge \\
\texttt{jrb239@cam.ac.uk} \\
\And
Yiren Zhao \\
Imperial College London and University of Cambridge\\
\texttt{a.zhao@imperial.ac.uk} \\
\AND
Ilia Shumailov \\
University of Oxford \\
\texttt{ilia.shumailov@chch.ox.ac.uk}
\And
Robert D Mullins \\
University of Cambridge \\
\texttt{robert.mullins@cl.cam.ac.uk}
}

\begin{document}

\maketitle

\begin{abstract}
    Given the wide and ever growing range of different efficient Transformer attention mechanisms, it is important to identify which attention is most effective when given a task. In this work, we are also interested in combining different attention types to build heterogeneous Transformers. We first propose a DARTS-like Neural Architecture Search (NAS) method to find the best attention for a given task, in this setup, all heads use the same attention (homogeneous models). Our results suggest that NAS is highly effective on this task, and it identifies the best attention mechanisms for IMDb byte level text classification and Listops. We then extend our framework to search for and build Transformers with multiple different attention types, and call them heterogeneous Transformers. We show that whilst these heterogeneous Transformers are better than the average homogeneous models, \emph{they cannot outperform the best}. We explore the reasons why heterogeneous attention makes sense, and why it ultimately fails.
\end{abstract}

\section{Introduction}

Since the first proposal of the Transformer architecture by \citet{tfm}, many alternatives have been proposed for the attention mechanism \citep{bigbird,linear_tfm,linformer,local_tfm,longformer,performer,reformer,sparse_tfm,synthesizer} due to the original dot-product attention mechanism having quadratic complexities in time and space with respect to the length of the input sequence.
Recent work \citep{lra} showed that these alternative architectures often perform well at different tasks when there is no pretraining, thus there is no clear attention that is best at every type of task.
Therefore we ask:
\emph{Can we efficiently learn the best attention for a given long range task?}

It is thought that each attention head in the Transformer can learn a different relationship, much like how in Convolutional Neural Networks (CNNs) each kernel learns a different feature.
\citet{lra} hypothesizes that each attention mechanism represents a different functional bias for which attention relationships should be learned, and that the utility of this bias is dependent on the task and its processing requirements.
Thus if we had many different attention types in a Transformer, they could each more easily learn different types of relationship, and thus make the Transformer more effective overall.
\emph{So is the optimal attention for a task a mixture of different attentions?}

\begin{figure}[t]
    \centering
    \scriptsize
    \includesvg[width=0.9\textwidth]{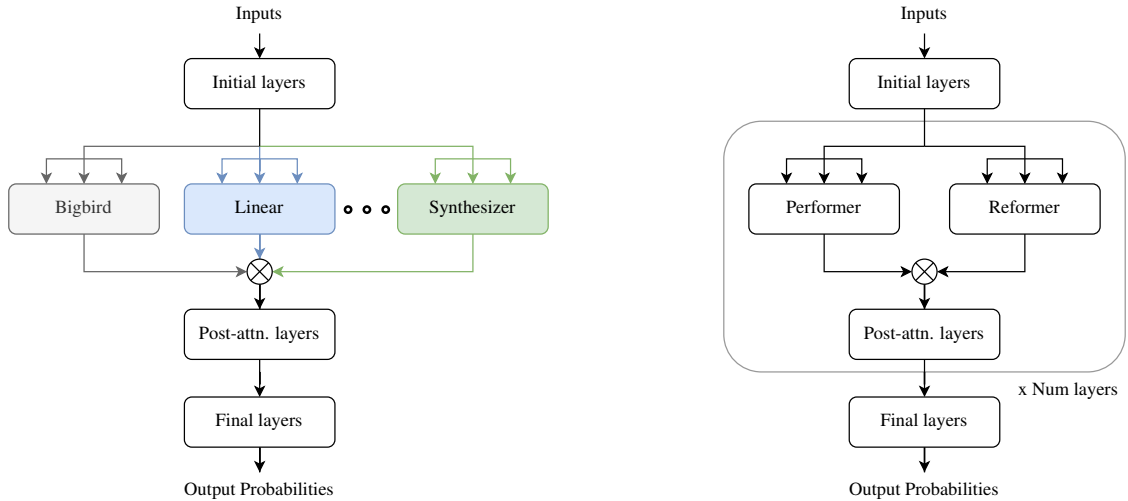}
    \caption{Our supernetwork architecture (left) that we search over, and an example final derived architecture (right) that is heterogeneous across heads and homogeneous across layers. The supernetwork is a single layer Transformer with several different multi-head attention blocks whose outputs are averaged.}
    \label{fig:strategy}
\end{figure}

In this paper we apply Neural Architecture Search (NAS) techniques to the Transformer attention search space and propose DARTFormer, a DARTS-like method \citep{darts} for finding the best attention, a high level illustration of the method is presented in \Cref{fig:strategy}.
To do this we use multiple candidate attention types in parallel and sum their outputs in the attention mechanism of the Transformer.
Following \citet{rethink_darts} we use masked validation accuracy drop as our metric for determining the performance of each attention type.

For clarity we refer to computation of the QKV linear projections, multi-head attention, and then concatenation and linear projection back down to the sequence features as the attention block.
In a standard Transformer there is only a single attention block that has multiple heads.
When training mixed attention models (either supernetworks or a final heterogeneous model) we use multiple attention blocks, each containing only a single attention type.

Following \citet{lra}, we use a representative mixture of different attention mechanisms as part of our search space in order to cover the main methods of achieving efficient attention.
The specific attentions we use are: Bigbird \citep{bigbird}, Linear Transformer \citep{linear_tfm}, Linformer \citep{linformer}, Local attention \citep{local_tfm}, Longformer \citep{longformer}, Performer \citep{performer}, Reformer \citep{reformer}, Sparse Transformer \citep{sparse_tfm}, and Synthesizer \citep{synthesizer}.

We use our setup to investigate two key paradigms.
The first is learning the best attention for a new task with a single layer Transformer, this means only one full-scale Transformer needs to be trained after a good attention mechanism is found.
The second paradigm, illustrated in \Cref{fig:strategy}, is using a single layer Transformer to find the best head-wise heterogeneous attention mixture for that task, and then using that mixture in each layer of a full Transformer model, making it layer-wise homogeneous.
We test these paradigms on three different tasks.
They are taken from \citet{lra} and were specifically designed to test the capabilities of efficient long range Transformers.

In this paper we make the following contributions:

\begin{itemize}
    \item We propose a DARTS-like framework to efficiently find the best attention for a task.
    \item We extend this framework to building and searching for optimal heterogeneous attention Transformer models.
    \item We empirically show that heterogeneous Transformers \emph{cannot} outperform the best homogeneous Transformer for our selected long range NLP tasks. 
\end{itemize}

\section{Related Work}

\subsection{Transformer Attention}\label{sec:tfm_attn}

Since \citet{tfm}, a large variety of replacement attention mechanisms were proposed.
These use different approaches, such as, low rank approximations and kernel based methods \citep{linformer, performer, linear_tfm, synthesizer}, fixed/factorised/random patterns \citep{synthesizer, sparse_tfm, bigbird, longformer, sinkhorn}, learnable patterns \citep{reformer, sinkhorn}, recurrence \citep{tfm_xl} and more \citep{set_tfm}.
\citet{efficient_tfms} gives a detailed survey of different attention mechanisms.

\citet{lra} compares the performance, speed and memory usage of many of these different attention mechanisms on a variety of tasks, including NLP and image processing.
Their main finding is that the performance of each attention mechanism is highly dependent on the nature of the task being learned when pretrained embeddings or weights aren't used.
This motivates the initial part of our research.

\subsection{NAS on Transformers}

Recent work  of \citet{tfm_nas_efficient} applies RankNAS \citep{rank_nas} to cosFormer \citep{cosformer} and standard Transformers \citep{tfm}.
They perform a search of the hyperparameters on a cosFormer network and compare these to the same method applied to the standard Transformer.
\citet{tfm_nas_one_shot} searches the hyperparameter space of a BERT \citep{bert} model architecture heterogeneously to find an optimal efficient network.
These methods search hyperparameters, whereas we fix hyperparameters and search over the space of attention mechanisms.

\section{Method}

\subsection{DARTS Style NAS}

We train a supernetwork that contains attention heads of each attention type.
This is analogous to the edges in a DARTS cell containing all possible operations for that edge.
We also use `fixed $\alpha$' with masked validation accuracy as our metric for the strength of each edge as detailed in \citet{rethink_darts}.
Standard DARTS assigns a weight for each edge in the computation graph and softmaxes this on the output of each cell (see \Cref{fig:darts_tfm}, left, in \Cref{appendix:sup_diag}), since we are using a 'fixed $\alpha$' approach we do not need to do this and simply average our edges without using learnable weights.
This allows us to train the supernetwork until convergence and then select the best attention (or prune out the worst). Note that this removes the need to carry out the bi-level optimisation that is required in the original DARTS paradigm that uses validation performance to train the edge weights.
Our approach is illustrated in \Cref{fig:strategy}.
The DARTS supernetwork cell and the standard Transformer architecture are given in \Cref{appendix:sup_diag} in \Cref{fig:darts_tfm}, to show how our architecture relates to them.

\subsection{Architecture and Search Space}

Our Transformer encoder supernetwork consists of an embedding layer, an attention block supernetwork, a feed-forward network (FFN), and a linear classifier layer.
In normal Transformer attention, each attention block consists of: QKV linear projections from features to number of heads $\times$ head dimension, scaled dot product attention on each head, concatenation of heads, and a final dense projection back to the feature dimension \citep{tfm}.
In theory, we want to search over the space of alternatives for the scaled dot product attention operation independently for each head.
This would replace scaled dot product attention in a normal Transformer with an average of candidate attention mechanisms in a DARTS-like paradigm.
The averages from each head would then be concatenated and linearly projected back to the feature dimension as normal.

However, some attention mechanisms, such as Reformer \citep{reformer}, modify the linear projections or concatenation operations.
Because of this we instead search over candidate multi-head attention blocks where each one implements the projections, attention, and concatenation operations.
This is illustrated in \Cref{fig:search} in \Cref{appendix:sup_diag} .
When the candidate blocks are single head and we have $H$ blocks per candidate attention mechanism, after the architecture search is complete, the computation graph is equivalent to if we searched over the attention mechanisms themselves with $H$ heads.
With the search at the block level with single heads, each attention mechanism can learn its own linear projections, whereas in a search over just the attention mechanisms these would be shared within each head.
The disadvantage of this is increased memory and computation during the search, since now heads do not share the linear projection.

\subsection{Experiments}\label{section:experiments}

\subsubsection{Finding Optimal Homogeneous Attention}

For the first experiment we train a single layer network with a block for each attention mechanism.
These blocks are each initialized with the desired number of final heads for that task.
After training we perform a single masked validation accuracy trial and pick the highest scoring mechanism as a good candidate mechanism for that task when used in a full Transformer model homogeneously.
This paradigm is summarized in an algorithmic form in \Cref{alg:exp1} in \Cref{appendix:algorithms}.

\subsubsection{Finding Optimal Heterogeneous Attention}

In this paradigm we try to learn an optimal single layer with a fixed number of heads and stack it, analogous to searching for an optimal cell in DARTS.

We begin with $H$ attention blocks of each attention type, each one a single head, where $H$ is the desired number of final attention heads in our heterogeneous Transformer.
We train the supernetwork to rough convergence and then alternate between removing the worst attention block and fine-tune training the remaining model.
We do this until we have only $H$ single-head attention blocks left, each potentially a different attention mechanism.

Once we have the optimal layer architecture for the task, we reinitialize the Transformer network with multiple layers where each one contains the mix of attentions found to be optimal.
Where possible we group attentions of the same type into the same attention block with multiple heads.
We then train this full Transformer model on the task until convergence.
This paradigm is summarized in \Cref{alg:exp2} in \Cref{appendix:algorithms}.
We will refer to this method as `NAS Prune'.

We also test a simpler method where we take the top 4 scoring attentions from the single layer to find the best homogeneous attention, and then train a full Transformer network with an even mix of these attentions at each layer.
This allows us to evaluate the general capabilities of heterogeneous attention.
We will refer to this method as `NAS One-shot'.

We also experiment with a head-wise homogeneous, layer-wise heterogeneous attention Transformer.
This is detailed in \Cref{appendix:lhet}.

\subsection{Attention Evaluation}

In order to evaluate the strength of each block we used a masked validation accuracy drop metric \citep{rethink_darts}.
First we get a baseline accuracy by running our model on the validation set.
Then, for each block, we mask it out and record the new validation accuracy.
This is illustrated in \Cref{fig:masked_val} in \Cref{appendix:sup_diag}.
The best block is the one which drops the accuracy the most, and the worst either drops it the least or increases it the most.

\section{Evaluation}

For each experiment we run the search on 3 different tasks taken from Long Range Arena (LRA) \citep{lra}.
These are: byte level binary text classification on IMDb dataset with 1k sequence length, Listops - 10-way classification with 2k sequence length, and byte level document matching which uses 4k sequence length.
Hyperparameters typically match those used in LRA, details are given in the \Cref{appendix:exp_details}.
One notable difference is that we train each homogeneous model 3 times and take an average.
The Listops and document matching tasks are also trained for longer to allow the models to better converge.

\subsection{Finding Optimal Homogeneous Attention}

\Cref{table:sgl_select_tc} shows us that our selection process is effective at identifying the best attention mechanism for the text classification and Listops tasks.
For both it also picks out the third best performing model with either its own second or third best.
For the document matching task, however, the single layer fails to find one of the top 3 attention mechanisms within top 3 picks.
One notable difference between its' document matching scores and the scores for the other two are that they are much smaller.
This could indicate that when the score is low magnitude, the model is not confident in its selection.
This can then be used to identify if it made a poor choice in advance of training the full homogeneous model.

\begin{table}[htb]
    \caption{Test accuracy for the full model based on results of \citet{lra}, and masked validation accuracy impact (more positive is better) for our trained single layer multiple attention block network (\textbf{first}, \underline{second}, \emph{third}).}
    \label{table:sgl_select_tc}
    \begin{center}
        \begin{tabular}{c | c c | c c | c c}
            \multirow{2}{*}{\bf Attention type} 
            & \multicolumn{2}{c|}{\textbf{Text classification}} & \multicolumn{2}{c|}{\textbf{Listops}} & \multicolumn{2}{c}{\textbf{Document Matching}} \\ 
            & Acc & Score & Acc & Score & Acc & Score \\
            \hline
            Bigbird            & \emph{62.7} & \emph{0.30} & 36.7 & 0.25 & 63.9 & \emph{0.011} \\
            Linear Transformer & \underline{64.4} & 0.17 & \underline{37.1} & -0.40 & 64.2 & -0.006 \\
            Linformer          & 58.6 & -0.28 & 29.9 & 0.00 & \emph{64.5} & 0.006 \\
            Local              & 56.3 & 0.14 & 36.8 & 0.30 & 58.0 & 0.000 \\
            Longformer         & 61.8 & -0.20 & 36.8 & \emph{0.40} & 61.2 & \textbf{0.039} \\
            Performer          & \textbf{64.5} & \textbf{0.51} & 36.2 & 0.30 & \underline{65.0} & 0.000 \\
            Reformer           & 55.7 & \underline{0.44} & \textbf{37.8} & \textbf{11.85} & 58.3 & \underline{0.028} \\
            Sparse Transformer & 61.3 & 0.02 & \emph{37.0} & \underline{0.50} & 63.2 & 0.000 \\
            Synthesizer        & 61.4 & -0.03 & 36.5 & 0.30 & \textbf{71.1} & -0.006 \\
        \end{tabular}
    \end{center}
\end{table}

\subsection{Finding Optimal Heterogeneous Attention}

Table \ref{table:head_het_layer_hom} shows the performance for the best homogeneous model, the fully trained model with an architecture chosen by the pruning NAS method (NAS Prune), and the fully trained model with an architecture chosen by taking the best four attentions from the single layer selection to find the best attention (NAS One-shot).

\begin{table}[htb]
    \caption{The selected attention mixes with accuracy of full model vs the accuracy of the best homogeneous models for each task}
    \label{table:head_het_layer_hom}
    \begin{center}
        \begin{tabular}{c | l l l l l l l }
            \multirow{2}{*}{\bf Task} & \multirow{2}{*}{$H$} & \multicolumn{2}{c}{\bf Best Homogeneous} & \multicolumn{2}{c}{\bf NAS Prune} & \multicolumn{2}{c}{\bf NAS One-shot} \\
            & & Attention & Acc & Attentions & Acc & Attentions & Acc \\
            \hline
            \multirow{5}{1.8cm}{\centering Text Classification} & \multirow{5}{*}{8}
                & \multirow{5}{*}{Performer} & \multirow{5}{*}{64.5}
                & Linear & \multirow{5}{*}{63.9} & \multirow{5}{2.2cm}{Bigbird x2 Linear x2 Performer x2 Reformer x2} & \multirow{5}{*}{64.4} \\
                & & & & Performer x2 & & \\
                & & & & Reformer x2 & & \\
                & & & & Sparse x2 & & \\
                & & & & Synthesizer & & \\
            \hline
            \multirow{5}{*}{Listops} & \multirow{5}{*}{8}
                & \multirow{5}{*}{Reformer} &\multirow{5}{*}{37.8}
                & Linear & \multirow{5}{*}{36.3} & \multirow{5}{2.2cm}{Local x2 Longformer x2 Reformer x2 Sparse x2} & \multirow{5}{*}{37.1} \\
                & & & & Longformer & & \\
                & & & & Performer & & \\
                & & & & Reformer x3 & & \\
                & & & & Sparse x2 & & \\
            \hline
            \multirow{4}{1.8cm}{\centering Document Matching} & \multirow{4}{*}{4}
                & \multirow{4}{*}{Synthesizer} & \multirow{4}{*}{71.1}
                & BigBird & \multirow{4}{*}{67.0} & \multirow{4}{2.2cm}{Bigbird Linformer Longformer Reformer} & \multirow{4}{*}{64.7} \\
                & & & & Linear & & \\
                & & & & Sparse & & \\
                & & & & Synthesizer & & \\
        \end{tabular}
    \end{center}
\end{table}

This shows us that heterogeneous Transformer networks fail to beat the best homogeneous model for any task.
We also experimented with choosing a mixture of the best performing homogeneous attentions, and having more attention heads of the better attention mechanisms for that task, but neither of these two variations gave consistent improvement on the NAS One-shot method.
This also shows us that the expensive pruning method poses no consistent advantages over the far cheaper NAS One-shot when it correctly identifies good attentions.

\section{Theories For Heterogeneous Attention Sub-optimality}

Whilst searching for the optimal homogeneous attention works well and validates our NAS method, our found heterogeneous attention Transformer models cannot beat the best homogeneous models.
This conflicts with the idea that the different mechanisms provide different biases which help the Transformer learn many different and useful relationships.
Here we provide some theories to explain what we have observed.

\emph{Different attention mechanisms do not bias the relationships learned.}
It might be that the relationships learned are not affected by the attention type, and the performance differences noted in \citet{lra} are due to some other effect. Thus the amount of functional bias by attention mechanisms is small.

\emph{Each tasks is best solved with attention that has learned specific types of relationships.} 
Whilst learning lots of relationships might be useful, perhaps each tasks requires a narrow subset of relationships to be learned, and thus a homogeneous attention well suited to this narrow subset is optimal.

\emph{Bad attention degrades performance.}
It might be that the different attention mechanisms do learn different relationships, and this generally helps, but a small number of poorly suited attention types degrade model performance.
This could be tested by more comprehensive studies on the effects of combining different attention types.

\emph{The training of other aspects of the Transformer is dependent on the attention mechanism.}
It might be that when trained in homogeneous models, the attention mechanism influences how other parts of the model learn, such as the feed-forward network.
Thus if there were multiple attention mechanisms, these parts might have conflicting pressures on their learning which cause them to not be well suited to any of the attention mechanisms.

\section{Conclusion}

For a given task it is often not clear which attention mechanism will perform best. Using a DARTS-like method you can train a single layer mixed attention Transformer and use masked validation accuracy drop to determine the best attention to use in a full Transformer model. If the masked validation drop scores are very low, the best attention may not have been correctly identified.

Heterogeneous attention Transformer networks typically perform better than the average homogeneous one. Reasonable combinations of attention can be found via a DARTS-like method. However, heterogeneous Transformers cannot beat a homogeneous model that is using the best attention mechanism for that given task.

\bibliography{paper}
\bibliographystyle{paper}

\appendix

\FloatBarrier

\section{Experimental Details}\label{appendix:exp_details}

In all tasks we tested on, we used an Adam optimizer with linear warm-up and square root decay. We used a base learning rate of 0.05, $\beta_1=0.9$, and $\beta_2=0.98$. We used a batch size of 32 for every task. All attention mechanisms used the [CLS] token for classification. Task specific hyperparameters are given in \cref{table:hyperparams}. In all Transformer models the feed-forward network (FFN) has a single hidden dimension. Before testing we take the model which had the best validation accuracy during training, the long train times are to ensure convergence.

\begin{table}[htb]
    \caption{Task specific hyperparameters, see introduction for definitions.}
    \label{table:hyperparams}
    \begin{center}
        \begin{tabular}{l | l l l l l l}
            \toprule
            {\bf Task} & \textbf{Training Steps} & \textbf{Warmup} & \textbf{Seq. Length} & $E$ & $A$ & $M$ \\
            \midrule
            IMDb Token Level & $30k$ & $8k$ & $1k$ & 512 & 64 & 2048 \\
            IMDb Byte Level & $30k$ & $8k$ & $1k$ & 512 & 64 & 2048 \\
            Listops & $10k$ & $1k$ & $2k$ & 512 & 64 & 2048 \\
            Document Matching & $10k$ & $8k$ & $4k$ & 128 & 32 & 512 \\
            \bottomrule
        \end{tabular}
    \end{center}
\end{table}

\section{Layer-wise Heterogeneous Attention}\label{appendix:lhet}

Here we try another method of heterogeneous attention.
Instead of using layers with an identical mixture of attentions at each layer, we make each layer homogeneous in its attention but use different attentions for different layers.

Given we want each layer to have $H$ attention heads of a single type, we initialize an attention block with $H$ heads for each candidate attention mechanism in each layer.
We train the supernetwork to rough convergence and then alternate between removing an attention block in a layer and fine-tune training the remaining model.
We go through each layer starting with the first, removing the worst attention block before moving on to the next layer.
Once we remove a block from the final layer we begin again at the first layer.
We keep passing through the network, removing and fine-tuning, until only one attention block remains in each layer.

We then take the attention blocks learned for each layer and train a new Transformer from scratch with those attentions at each layer.
This paradigm is summarized in algorithm form in the appendix as \Cref{alg:exp3}.

Due to the high memory requirements of this task, we use block sampling when training.
This means that for each training batch we only use a randomly selected subset of the candidate blocks in each layer.
This reduces memory requirements as there are less intermediate values and computations, but can affect the training dynamics.
Using this method within a DARTS like paradigm is shown to work well by \citet{pc_darts}.

We test this method on the text classification task which has 1k input sequence length to further limit computation requirements.
As the Transformers used in LRA for this task had 6 layers, we searched for a 6 layer model with layer-wise attention heterogeneity. The resulting architecture was:

Performer $\rightarrow$ Bigbird $\rightarrow$ Performer $\rightarrow$ Sparse $\rightarrow$ Performer $\rightarrow$ Sparse

When trained with the same hyperparameters as the LRA models, it achieved a test accuracy of 64.6\%. This puts this model around average for this task, unable to beat several attentions such as the Linear Transformer which had a test accuracy of 65.9\%, or the Performer which achieved 65.4\%.

\section{Supplementary Diagrams}\label{appendix:sup_diag}

\begin{figure}[hbt]
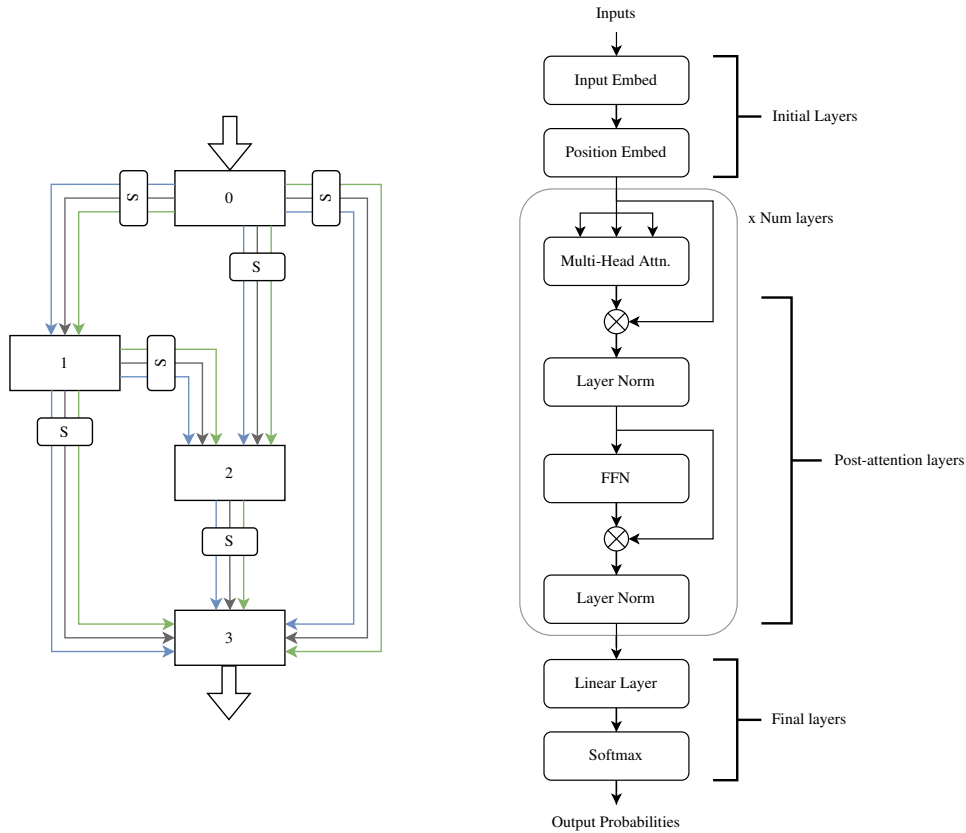

    \centering
    \begin{subfigure}{.3\textwidth}
        \tiny
        \includesvg[width=\textwidth]{imgs/darts_cell.svg}
    \end{subfigure}
    \hspace{.1\textwidth}
    \begin{subfigure}{.37\textwidth}
        \tiny
        \includesvg[width=\textwidth]{imgs/transformer.svg}
    \end{subfigure}
    \caption{DARTS Cell (left) and typical Transformer architecture (right). In DARTS inputs to all blocks are summed and 'S' denotes a softmax layer. Also shown is how the layers in the Transformer relate to the diagram in figure \ref{fig:strategy}.}
    \label{fig:darts_tfm}
\end{figure}

\begin{figure}[hbt]
    \centering
    \tiny
    \includesvg[width=\textwidth]{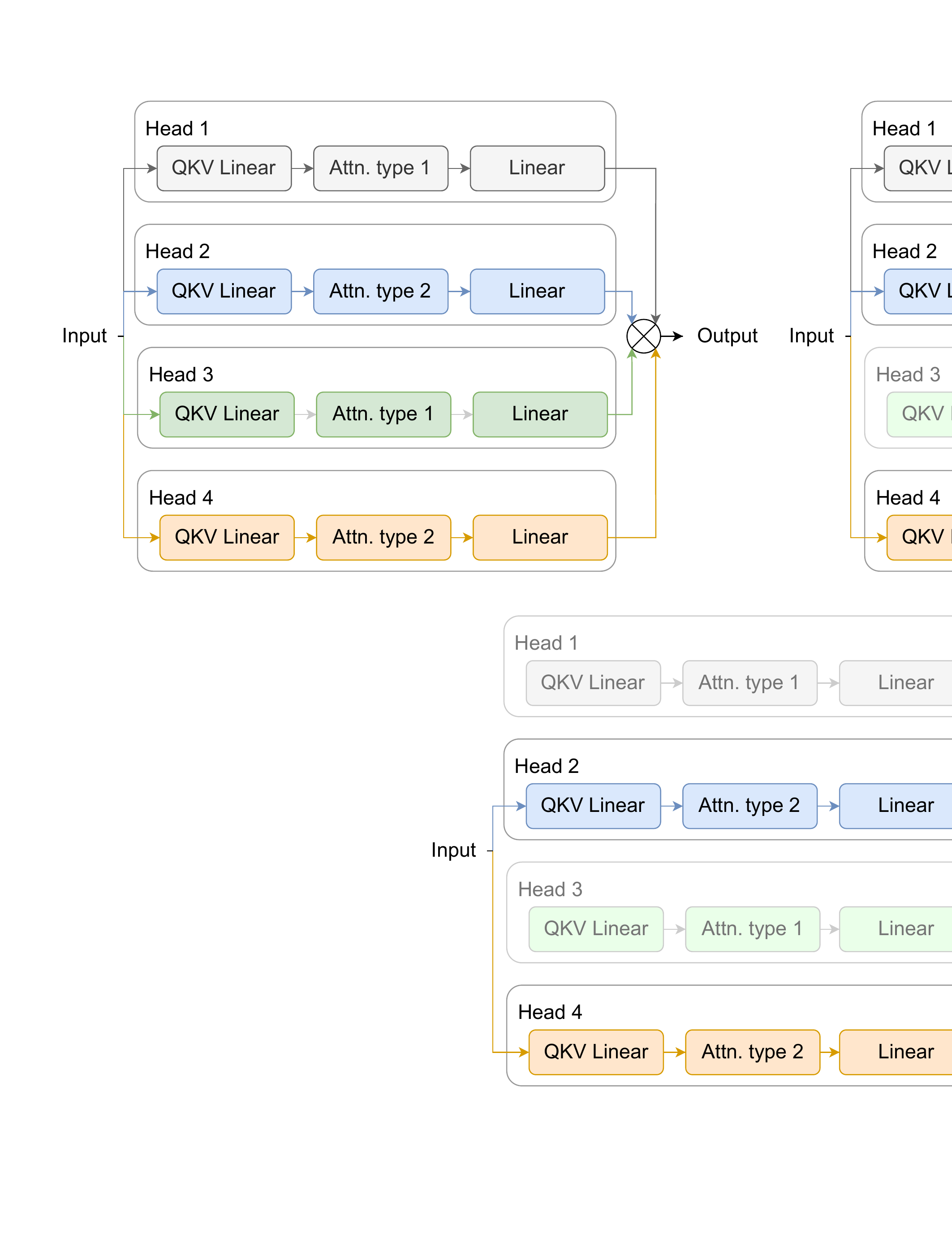}
    \caption{From left to right and top to bottom this figure shows how we remove from the model the worst attention block until we are left with just the best ones.}
    \label{fig:masked_val}
\end{figure}

\begin{figure}[hbt]
    \centering
    \begin{subfigure}{0.6\textwidth}
        \tiny
        \includesvg[width=\textwidth]{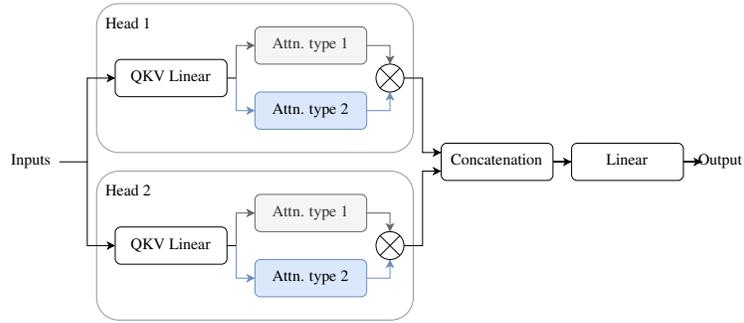}
        \caption{Theoretical multi head attention supernetwork}
    \end{subfigure}

    \vspace{0.5cm}
    \begin{subfigure}{0.6\textwidth}
        \tiny
        \includesvg[width=\textwidth]{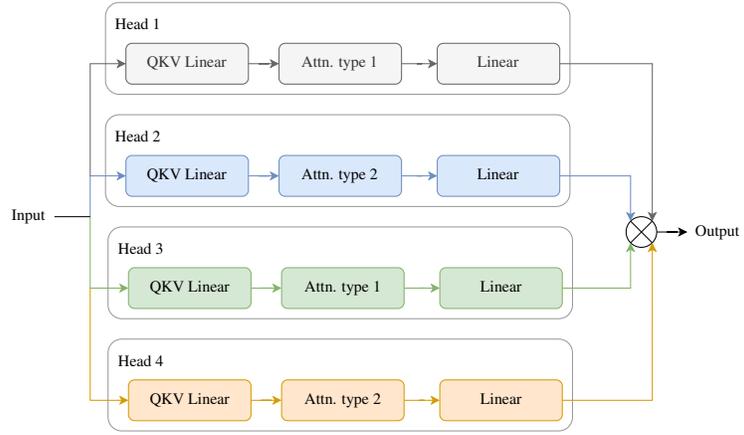}
        \caption{Actual multi head attention supernetwork}
    \end{subfigure}

    \vspace{0.5cm}
    \begin{subfigure}{0.6\textwidth}
        \tiny
        \includesvg[width=\textwidth]{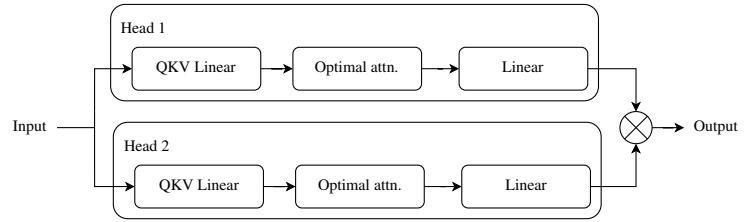}
        \caption{Final multi attention layer after architecture search}
    \end{subfigure}
    \caption{Diagrams of the multi head attention layer with the theoretical attention search space and the actual one used. Also shown is the resulting multi head attention architecture which would be equivalent in both cases.}
    \label{fig:search}
\end{figure}

\Cref{fig:darts_tfm,fig:masked_val,fig:search} help illustrate how our approach relates to DARTS and the original Transformer, attention heads being pruned during the search process in finding optimal heterogeneous attention, and an overview of the differences between the theoretical and actual supernetwork.

\FloatBarrier

\section{Algorithms}\label{appendix:algorithms}

Algorithms 1 and 2 correspond to the paradigms outlined in \cref{section:experiments}.

Let $\mathbb{B}_i$ denote the set of attention blocks in layer $i$, with $\mathbb{L}$ denoting the set of layers. We also use $a$ to represent accuracy and $s$ to represent a score with $\mathbb{S}_i$ representing the set of scores for the attention blocks in layer $i$. For single layer networks, the layer subscript is omitted.

\begin{algorithm}[hbt]
    \caption{Finding optimal homogeneous attention}
    \label{alg:exp1}
    \begin{algorithmic}[1]
        \State Initialise a single layer supernetwork with a multi-head attention block for each attention type
        \State Train supernetwork until convergence
        \State $a_{base} \gets$ Validation accuracy of supernetwork
        \For{$b_i \in \mathbb{B}$}
            \State Mask out $b_i$
            \State $a_i \gets$ Validation accuracy
            \State $s_i \gets a_{base} - a_i$
        \EndFor
        \State Selected attention mechanism $\gets \max(\mathbb{S})$ 
        \State Initialise full Transformer using the selected attention mechanism
        \State Train until convergence
    \end{algorithmic}
\end{algorithm}

\begin{algorithm}[hbt]
    \caption{Finding optimal heterogeneous attention}
    \label{alg:exp2}
    \begin{algorithmic}[1]
        \State Initialise a single layer super network with $H$ single-head attention blocks for each attention type, where $H$ is the desired number of heads per layer in the final network.
        \State Train supernetwork until rough convergence
        \Repeat
            \State $a_{base} \gets$ Validation accuracy of supernetwork
            \For{$b_i \in \mathbb{B}$}
                \State Mask out $b_i$
                \State $a_i \gets$ Validation accuracy
                \State $s_i \gets a_{base} - a_i$
            \EndFor
            \If{$|\mathbb{B}| > H$}
                \State From $\mathbb{B}$ remove block corresponding to $\min(\mathbb{S})$
                \State Train remaining supernetwork a small amount
            \EndIf
        \Until{$|\mathbb{B}| = H$}
        \State $\mathbb{B}_{optimal} \gets \mathbb{B}$
        \State Initialise a new network with multiple layers, where each layer has blocks $\mathbb{B}_{optimal}$ 
        \State Train until convergence
    \end{algorithmic}
\end{algorithm}

\begin{algorithm}[hbt]
    \caption{Head-wise homogeneous, layer-wise heterogeneous mixed attention search}
    \label{alg:exp3}
    \begin{algorithmic}[1]
        \State Initialise a multi layer supernetwork with multi-head attention block for each attention type in each layer
        \State Train supernetwork until rough convergence
        \Repeat
            \For{$i \in \mathbb{L}$}
                \State $a_{base} \gets$ Validation accuracy of supernetwork
                \For{$b_{ij} \in \mathbb{B}_i$}
                    \State Mask out $b_{ij}$
                    \State $a_{ij} \gets$ Validation accuracy
                    \State $s_{ij} \gets a_{base} - a_{ij}$
                \EndFor
                \If{$|\mathbb{B}_i| > 1$}
                    \State From $\mathbb{B}_i$ remove block corresponding to $\min(\mathbb{S}_i)$
                    \State Train remaining supernetwork a small amount
                \EndIf
            \EndFor
        \Until{$|\mathbb{B}_i| = 1, \forall i \in \mathbb{L}$}
        \State Initialise a full Transformer model where the attention mechanism used in layer $i$ corresponds to the type of the last block left in $\mathbb{B}_i$
        \State Train until convergence
    \end{algorithmic}
\end{algorithm}

\FloatBarrier

\end{document}